%% file: main.tex
\documentclass[10pt,conference]{IEEEtran}

\newif\iffinal
\finaltrue

\newif\ifarxiv
\arxivtrue

\newif\ifieee
\ieeefalse

\IEEEoverridecommandlockouts
\usepackage{cite}
\usepackage{amsmath,amssymb,amsfonts}
\usepackage{algorithmic}
\usepackage{graphicx}
\usepackage{textcomp}
\usepackage{xcolor}

\usepackage[normalem]{ulem} %

\usepackage{url}

\usepackage{icomma} %
\usepackage[acronym]{glossaries} %
\usepackage{xspace}
\usepackage{booktabs}
\usepackage{arydshln}

\usepackage[dvipsnames,table]{xcolor}
\usepackage{balance}

\usepackage{multirow}

\newboolean{showcomments}
\setboolean{showcomments}{true}     %

\def\BibTeX{{\rm B\kern-.05em{\sc i\kern-.025em b}\kern-.08em
    T\kern-.1667em\lower.7ex\hbox{E}\kern-.125emX}}

\usepackage[caption=false,farskip=0pt]{subfig}

\ifieee
\else
\usepackage[colorlinks=true,allcolors=black]{hyperref}
\fi

\usepackage[capitalise]{cleveref} %
\usepackage{makecell}

\newcommand*{\RL}[2][]{\textcolor{Rhodamine}{[\textbf{\ifthenelse{\equal{#1}{}}{RL}{RL(#1)}}: #2]}}

\newcommand*{\LW}[2][]{\textcolor{teal}{[\textbf{\ifthenelse{\equal{#1}{}}{LW}{LW(#1)}}: #2]}}

\newcommand*{\DM}[2][]{\textcolor{red}{[\textbf{\ifthenelse{\equal{#1}{}}{DM}{DM(#1)}}: #2]}}

\newcommand*{\EN}[2][]{\textcolor{blue}{[\textbf{\ifthenelse{\equal{#1}{}}{DM}{EN(#1)}}: #2]}}

\newacronymstyle{long-short-br}
{%
  \GlsUseAcrEntryDispStyle{long-short}%
}%
{%
  \GlsUseAcrStyleDefs{long-short}%
}
\setacronymstyle{long-short-br}

\usepackage{transparent}
\usepackage{tikz}
\ifarxiv
    \newcommand\copyrighttext{%
  \scriptsize Accepted at IJCNN 2026. The final published version is available on IEEE Xplore (DOI: \href{https://doi.org/10.1109/IJCNN69854.2026.11729979}{\textcolor{blue}{10.1109/IJCNN69854.2026.11729979}}).}
    \newcommand\copyrightnotice{%
    \begin{tikzpicture}[remember picture,overlay]
    \node[anchor=south,yshift=30pt,xshift=0pt] at (current page.south) {\fbox{\transparent{0.85}\parbox{\dimexpr0.72\textwidth-\fboxsep-\fboxrule\relax}{\copyrighttext}}};
    \end{tikzpicture}%
    }
\else
\fi

\begin{document}

\input{0-acronyms}
\input{0-variables}

\title{\dataset: An Expanded Dataset for Fine-Grained License Plate Legibility Classification} %

\iffinal
\author{Lucas Wojcik\IEEEauthorrefmark{1}, Eduardo A. F. Machoski\IEEEauthorrefmark{1}, Eduil Nascimento Jr.\IEEEauthorrefmark{2}, Rayson Laroca\IEEEauthorrefmark{3}\textsuperscript{,}\IEEEauthorrefmark{1}, and David Menotti\IEEEauthorrefmark{1}\\[5pt]
\IEEEauthorblockA{\IEEEauthorrefmark{1}\hspace{0.15mm}Federal University of Paran\'a, Curitiba, Brazil}
\IEEEauthorblockA{\IEEEauthorrefmark{2}\hspace{0.15mm}Department of Technological Development and Quality, Paran\'{a} Military Police, Curitiba, Brazil}
\IEEEauthorblockA{\IEEEauthorrefmark{3}\hspace{0.15mm}Pontifical Catholic University of Paran\'a, Curitiba, Brazil}
\IEEEauthorrefmark{1}{\hspace{-0.2mm}\tt\small\{lmlwojcik, eafm23, menotti\}@inf.ufpr.br} \quad \IEEEauthorrefmark{2}{\hspace{0.15mm}\tt\small eduiljunior@pm.pr.gov.br} \quad \IEEEauthorrefmark{3}{\hspace{0.15mm}\tt\small \texttt{rayson@ppgia.pucpr.br}}
}
\fi

\maketitle

\ifarxiv
    \copyrightnotice
\else
\fi

\ifarxiv
  \vspace{-3mm}
\else
\fi

\input{0-abstract}


\input{1-intro}
\input{2-related}
\input{3-proposed}

\input{4-setup}
\input{5-results}
\input{6-conclusions}
\input{0-acknowledgments}

\balance

\bibliographystyle{IEEEtran}
\bibliography{bibliography}

\end{document}

%% file: 0-acronyms.tex
\newacronym{alpr}{ALPR}{Automatic License Plate Recognition}
\newacronym{gan}{GAN}{Generative Adversarial Network}
\newacronym{lp}{LP}{license plate}
\newacronym{lplcv1}{previous benchmark}{[HIDDEN FOR REVIEW]}
\newacronym{dataset}{novel benchmark}{HIDDEN}
\newacronym{ocr}{OCR}{Optical Character Recognition}
\newacronym{sr}{SR}{super-resolution}
\newacronym{ema}{EMA}{Exponential Moving Average}
\newacronym{senatran}{SENATRAN}{Brazilian National Traffic Secretariat}

%% file: 0-variables.tex
\iffinal
    \newcommand{\previous}{LPLCv1\xspace}
    \newcommand{\dataset}{LPLCv2\xspace}
\fi
\newcommand{\parseq}{PARSeq-tiny\xspace}
\newcommand{\ocrchina}{GP\_LPR\xspace}

\newcommand{\StateName}{%
  \iffinal
    Paraná\xspace
  \fi
}

\newcommand{\urlDataset}{%
  \iffinal
    \url{https://github.com/lmlwojcik/LPLCv2-Dataset}\xspace
  \fi
}

%% file: 0-abstract.tex
\ifarxiv
  \vspace{-1.9mm}
\else
\fi

\begin{abstract}

Modern \gls*{alpr} systems achieve outstanding performance in controlled, well-defined scenarios.
However, large-scale real-world usage remains challenging due to low-quality imaging devices, compression artifacts, and suboptimal camera installation.
Identifying illegible \glspl*{lp} has recently become feasible through a dedicated benchmark; however, its impact has been limited by its small size and annotation errors.
In this work, we expand the original benchmark to over three times the size with two extra capture days, revise its annotations and introduce novel labels.
\gls*{lp}-level annotations include bounding boxes, text, and legibility level, while vehicle-level annotations comprise make, model, type, and color.
Image-level annotations feature camera identity, capture conditions (e.g., rain and faulty cameras), acquisition time, and day ID.
We present a novel training procedure featuring an Exponential Moving Average-based loss function and a refined learning rate scheduler, addressing common mistakes in testing.
These improvements enable a baseline model to achieve an 89.5\%~F1-score on the test set, considerably surpassing the previous state of the art.
We further introduce a novel protocol to explicitly addresses camera contamination between training and evaluation splits, where results show a small impact.
Dataset and code are publicly available at \urlDataset.

\end{abstract}

%% file: 1-intro.tex
\section{Introduction}

\glsresetall

\gls*{alpr} systems have become widely adopted in recent years, supporting applications such as road surveillance, law enforcement, and parking management~\cite{ismail2025automatic}.
Advances in deep learning have enabled \gls*{alpr} systems to achieve remarkably high performance in core tasks, including \gls*{lp} detection and \gls*{lp} text recognition~\cite{laroca2025advancing,ke2023ultra,ding2024endtoend}, particularly under controlled~conditions.

Several datasets commonly used in \gls*{alpr} research suffer from inherent limitations, including a small number of capture devices~\cite{goncalves2016ssig}, often producing high-quality images not representative of real-world surveillance conditions, limited diversity in acquisition scenarios~\cite{englishlp}, and relatively small sample sizes~\cite{clpd}.
When deployed under challenging conditions such as rain, fog, low-light environments, or in the presence of low-resolution imagery, suboptimal camera placement, inadequate acquisition equipment, and compression artifacts, the performance of state-of-the-art \gls*{alpr} systems degrades significantly~\cite{wahyu2024fog,nascimento2025toward}.
These shortcomings underscore the need for more diverse, realistic, and challenging \gls*{alpr}~datasets.

\begin{figure}[!t]
    \centering
    \captionsetup[subfigure]{labelformat=empty,captionskip=1.75pt,font=footnotesize}

    \subfloat[][]{\includegraphics[width=0.23\linewidth]{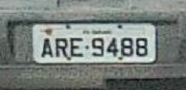}
    }
    \hspace{-1.5mm}
    \subfloat[][]{\includegraphics[width=0.23\linewidth]{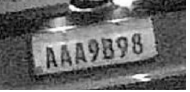}
    }
    \hspace{-1.5mm}
    \subfloat[][]{\includegraphics[width=0.23\linewidth]{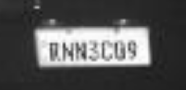}
    }
    \hspace{-1.5mm}
    \subfloat[][]{\includegraphics[width=0.23\linewidth]{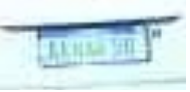}
    }

    \vspace{-3.3mm}

    \subfloat[][]{\includegraphics[width=0.23\linewidth]{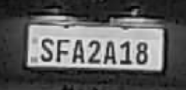}
    }
    \hspace{-1.5mm}
    \subfloat[][]{\includegraphics[width=0.23\linewidth]{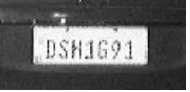}
    }
    \hspace{-1.5mm}
    \subfloat[][]{\includegraphics[width=0.23\linewidth]{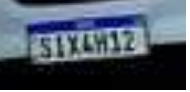}
    }
    \hspace{-1.5mm}
    \subfloat[][]{\includegraphics[width=0.23\linewidth]{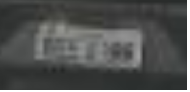}
    }

    \vspace{-3.3mm}

    \subfloat[][Perfect]{\includegraphics[width=0.23\linewidth]{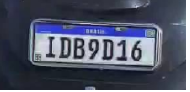}
    }
    \hspace{-1.5mm}
    \subfloat[][Good]{\includegraphics[width=0.23\linewidth]{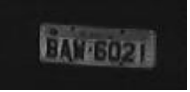}
    }
    \hspace{-1.5mm}
    \subfloat[][Poor]{\includegraphics[width=0.23\linewidth]{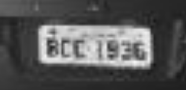}
    }
    \hspace{-1.5mm}
    \subfloat[][Illegible]{\includegraphics[width=0.23\linewidth]{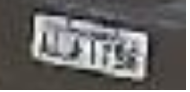}
    }
    
    \caption{\glspl*{lp} grouped by legibility class, following the definition of~\cite{wojcik2025lplc}.}
    \label{fig:lp_leg}
    \vspace{-1mm}
\end{figure}

At the same time, real-world ALPR applications demand fast and reliable license plate tracking and recognition under unconstrained conditions. Critical systems, such as those used for road surveillance, depend on low false positive rates to ensure efficient resource allocation and operational trustworthiness. Incorrect predictions can lead to unnecessary use of resources and raise ethical concerns.
As these issues are particularly pronounced in scenarios with low \gls*{lp} legibility, several strategies have been proposed to mitigate their impact.

\Gls*{sr} is one such example.
Despite recent advances in \gls*{sr} techniques aimed at recovering characters from severely degraded \glspl*{lp}~\cite{nascimento2024enhancing,pan2024lpsrgan}, this line of research remains incipient and is not yet reliable enough for deployment in real-world, safety-critical systems.
A key limitation is the limited cross-dataset generalization of existing methods.
As reported in~\cite{wojcik2025lplc}, current state-of-the-art license plate \gls*{sr} approaches not only struggle to generalize across datasets but can also harm high-quality inputs, reducing \gls*{lp} legibility and ultimately degrading \gls*{ocr}~performance.

Another example is \gls*{lp} legibility classification, a recently explored task that aims to distinguish between legible and illegible \glspl*{lp} to handle each case appropriately.
In large-scale real-world deployments, ALPR systems may process millions of images per day, making it computationally infeasible to apply end-to-end processing pipelines, such as \gls*{sr} followed by OCR, to every single image\footnote{This large-scale setting, involving millions of images per day, corresponds exactly to the operational scenario of one of our ongoing research projects.}.
Instead, legibility classification enables early decision-making, allowing systems to discard unusable samples, selectively apply \gls*{sr} to severely degraded \glspl*{lp}, or directly perform \gls*{ocr} on high-quality inputs.
Despite its practical relevance, prior work has reported accuracies of only around 76\% in the most general evaluation setting, highlighting substantial room for improvement~\cite{wojcik2025lplc}.

This work addresses key limitations of previous studies on legibility classification.
We revisit an existing benchmark~\cite{wojcik2025lplc} by correcting annotation errors, introducing new and more detailed labels, and extending it with many more images that cover a broader range of real-world scenarios.
The \gls{lp} legibility is annotated using four classes (see \cref{fig:lp_leg}).
In addition, we propose an improved training procedure tailored to the most common error patterns observed in baseline experiments. Our main contributions are summarized as follows:

\begin{itemize}
    \item A comprehensive revision of a previous benchmark, including the correction of mis-annotated legibility labels and its extension into a new dataset that is over three times larger than the original, with samples collected from more than $700$ cameras and covering scenarios underrepresented in the literature, such as faulty cameras, low-resolution imagery, and adverse weather~conditions;
    \item The introduction of novel, fine-grained annotations at both the image and \glspl*{lp} levels, including camera identifiers and capture conditions for images, as well as vehicle-related data for the \glspl*{lp};
    \item An improved training procedure specifically designed to mitigate common errors and better handle hard samples present in the dataset;
    \item A cross-camera evaluation to investigate potential camera contamination between training and testing partitions.
\end{itemize}

The remainder of this work is organized as follows.
\cref{sec:related} provides an overview of related methods and datasets.
\cref{sec:dataset} describes the proposed dataset and the revisions made to the previous benchmark.
\cref{sec:experiments} details the experimental protocol and setup.
\cref{sec:results} presents and discusses the results.
Finally, \cref{sec:conclusion} concludes the paper.

%% file: 2-related.tex
\section{Related Work}
\label{sec:related}

Recent advances in \gls{alpr} have led to very high detection and recognition rates on most public benchmarks~\cite{laroca2023leveraging,laroca2025advancing,liu2024improving}.
Modern pipelines are able to handle a wide range of conditions within these datasets, achieving strong performance on mainstream \gls{alpr} evaluation protocols.

A key challenge for cross-dataset generalization stems from the diversity of \gls{lp} layouts across countries. To address this issue, recent works have explored synthetic data generation as a means to increase layout coverage and robustness~\cite{laroca2025advancing}.
This diversity has also driven a shift from character-based \gls*{ocr} approaches~\cite{laroca2018robust,silva2020realtime} toward global recognition models~\cite{alzahrani2025attention,seo2022layout,liu2024irregular}.
Global models typically rely on implicit character localization, often through attention mechanisms, enabling improved recognition across different \gls*{lp}~layouts.

Despite these advances, most publicly available datasets remain biased toward high-quality, fully legible \glspl*{lp}.
For Brazilian and Mercosur layouts, commonly used datasets include SSIG-SegPlate~\cite{goncalves2016ssig}, UFPR-ALPR~\cite{laroca2018robust}, and RodoSol-ALPR~\cite{laroca2022cross}. RodoSol-ALPR is collected in relatively controlled toll plaza environments along a single highway, while SSIG-SegPlate and UFPR-ALPR rely on a very limited number of cameras (one and three, respectively) and predominantly feature high-resolution, clearly legible~\glspl*{lp}.

Chinese \glspl*{lp} are primarily represented by CCPD~\cite{xu2018towards} and CLPD~\cite{clpd}. Although CCPD is widely adopted, it has undergone multiple revisions and expansions, leading to inconsistencies in dataset size and evaluation protocols across studies~\cite{gao2023group,ding2024endtoend}. Differences in test splits across versions hinder direct comparison of reported results~\cite{laroca2025advancing}, and the presence of near-duplicate samples, corresponding to images of the same vehicle/\gls*{lp}, across training and test sets further compromises fair evaluation~\cite{laroca2023do}.
CLPD introduces challenges such as skewed \glspl*{lp}, varying illumination, and adverse weather conditions, as well as diverse capture angles and devices. However, image quality variability remains limited, with characters remaining clearly distinguishable despite geometric distortions.

Additional datasets include AOLP~\cite{hsu2013aolp}, CD-HARD~\cite{silva2018cdhard}, and OpenALPR-EU~\cite{openalpreu}. AOLP contains images captured with both handheld and static cameras, but is restricted to close-range, high-quality views. CD-HARD and OpenALPR-EU are comparatively small, with just over $100$ samples each, and also rely on closeup imagery acquired with high-quality cameras. While CD-HARD introduces severe viewpoint distortions, these datasets do not capture other common real-world degradations such as low resolution and severe blur.

Overall, existing datasets span multiple region formats but remain constrained by controlled capture conditions and high-resolution imagery. In unconstrained surveillance environments, \gls{alpr} systems must operate across diverse cameras, resolutions, and adverse conditions.
As a result, the practical applicability of current state-of-the-art methods remains limited in many real-world deployments.

To address these limitations, we introduce the \dataset dataset, described in detail in \cref{sec:dataset}.
This benchmark incorporates a legibility measure that explicitly quantifies OCR difficulty, enabling systematic evaluation under challenging conditions. 
The data was collected from over $700$ cameras spanning a wide range of resolutions, viewpoints, and capture distances, resulting in \glspl*{lp} with varying levels of degradation. In addition, we provide fine-grained annotations that allow \dataset to serve as a challenging benchmark for conventional and robust \gls{alpr} tasks.

%% file: 3-proposed.tex
\begin{figure*}[!htb]
\centering

\resizebox{0.95\linewidth}{!}{
\includegraphics[height=0.125\textwidth]{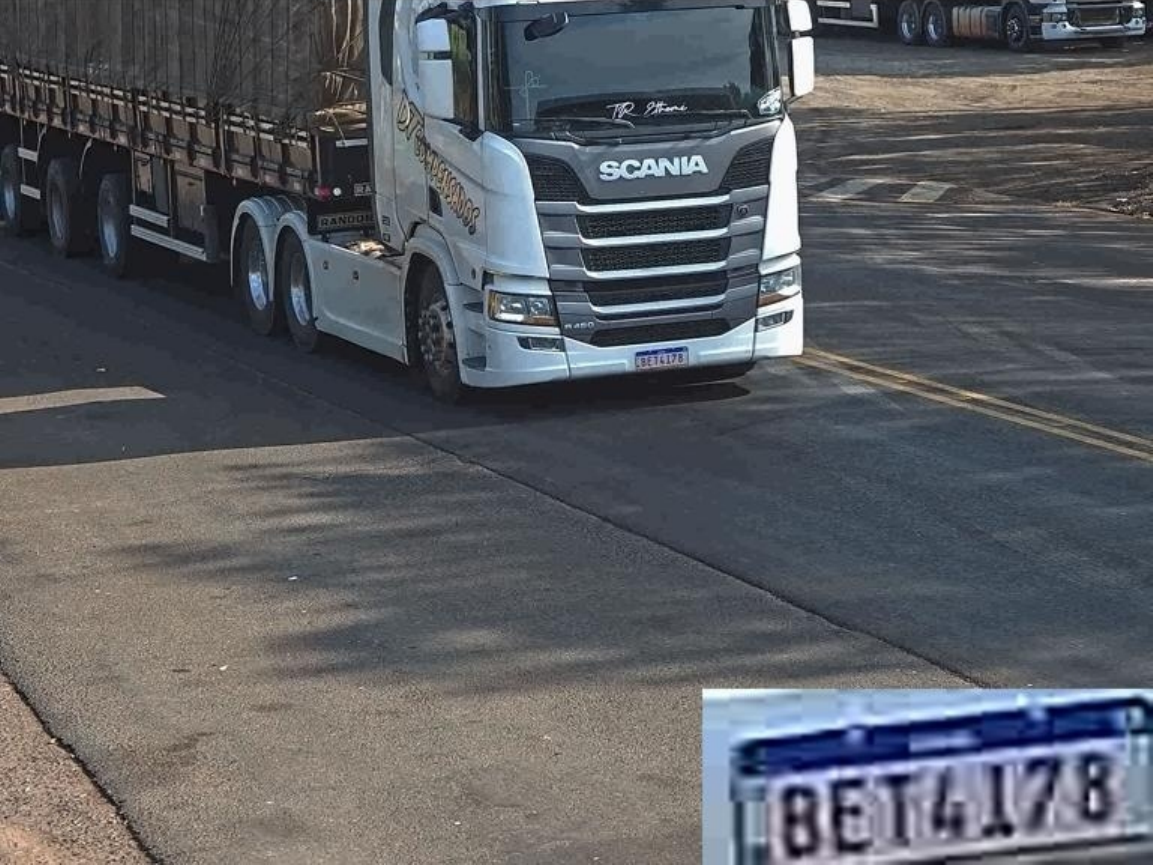}
\includegraphics[height=0.125\textwidth]{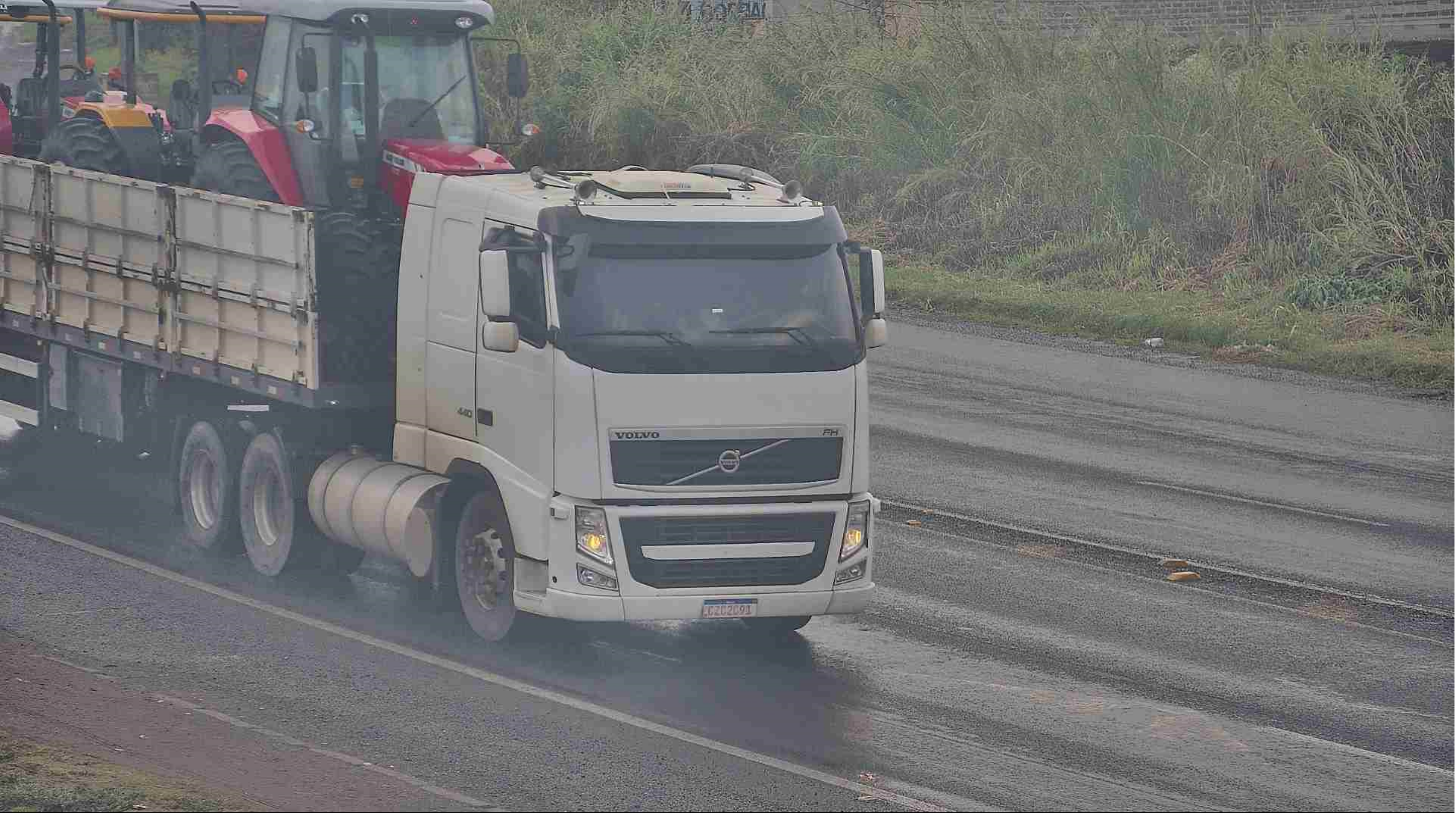}
\includegraphics[height=0.125\textwidth]{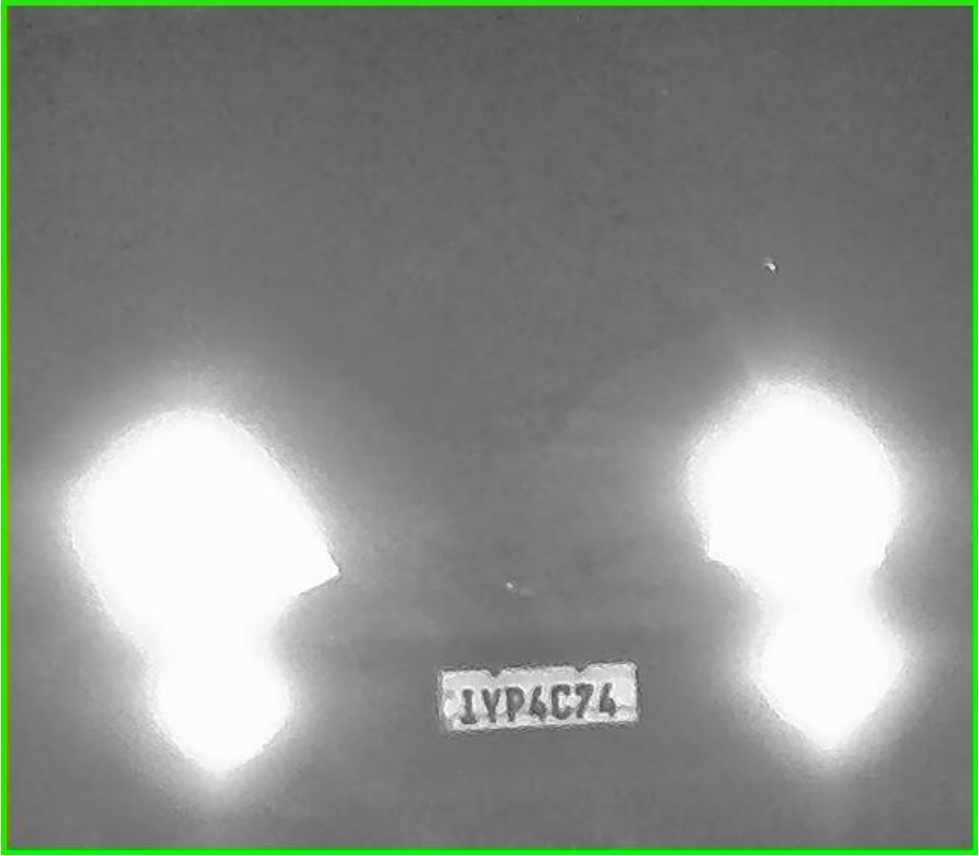}
\includegraphics[height=0.125\textwidth]{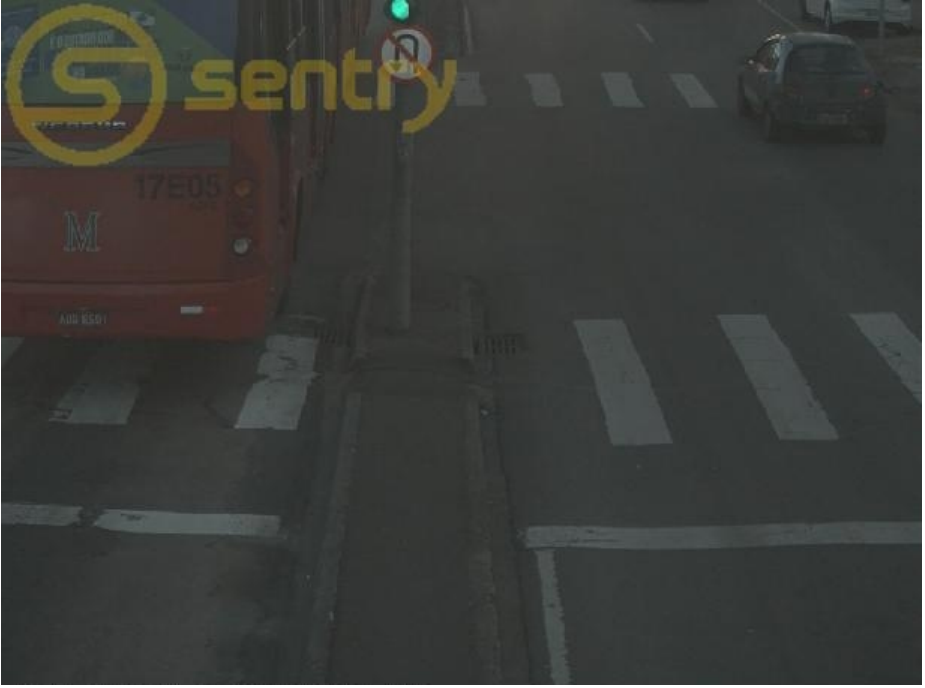}
\includegraphics[height=0.125\textwidth]{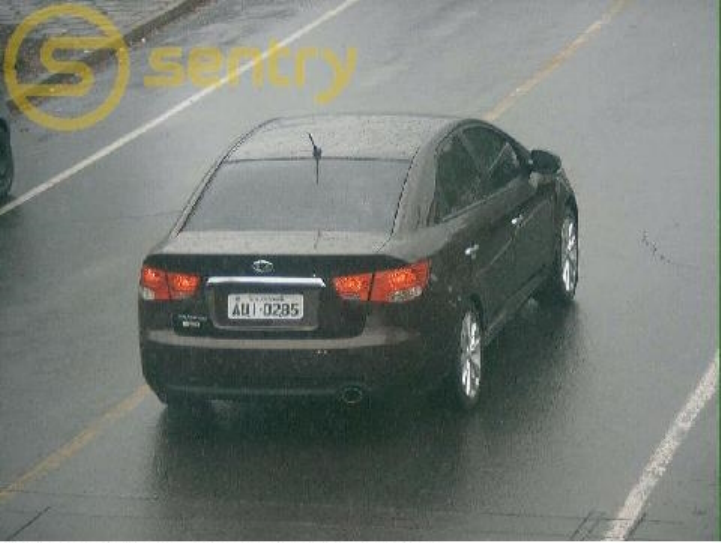}
}

\vspace{1.1mm}

\resizebox{0.95\linewidth}{!}{
\includegraphics[height=0.1175\textwidth]{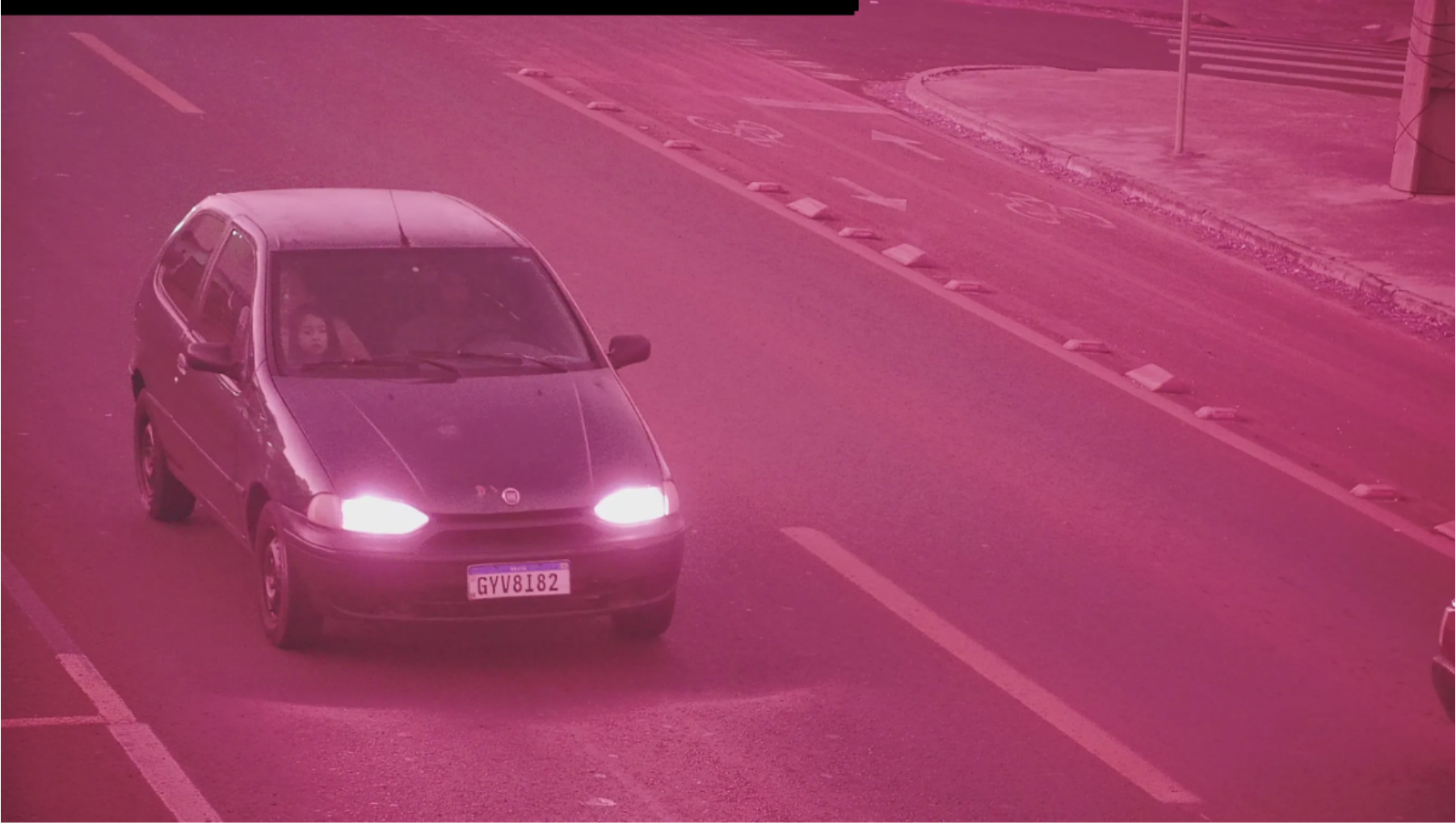}
\includegraphics[height=0.1175\textwidth]{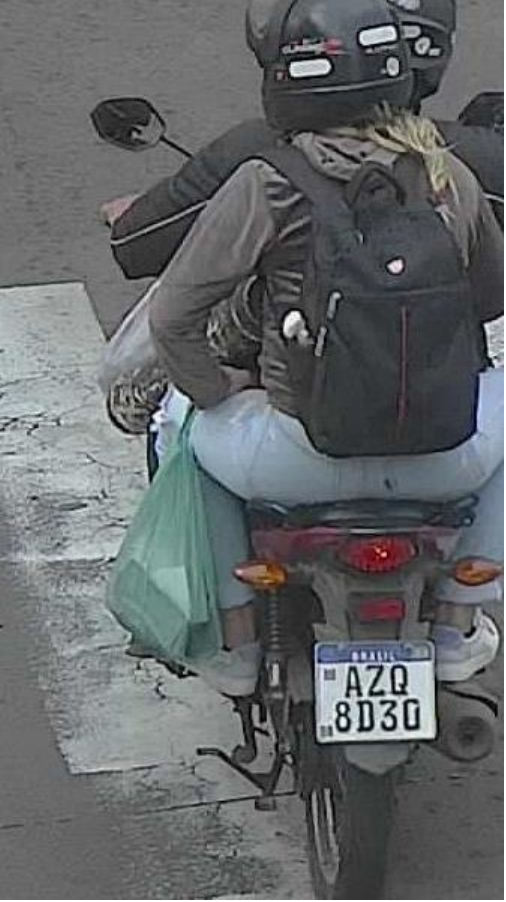}
\includegraphics[height=0.1175\textwidth]{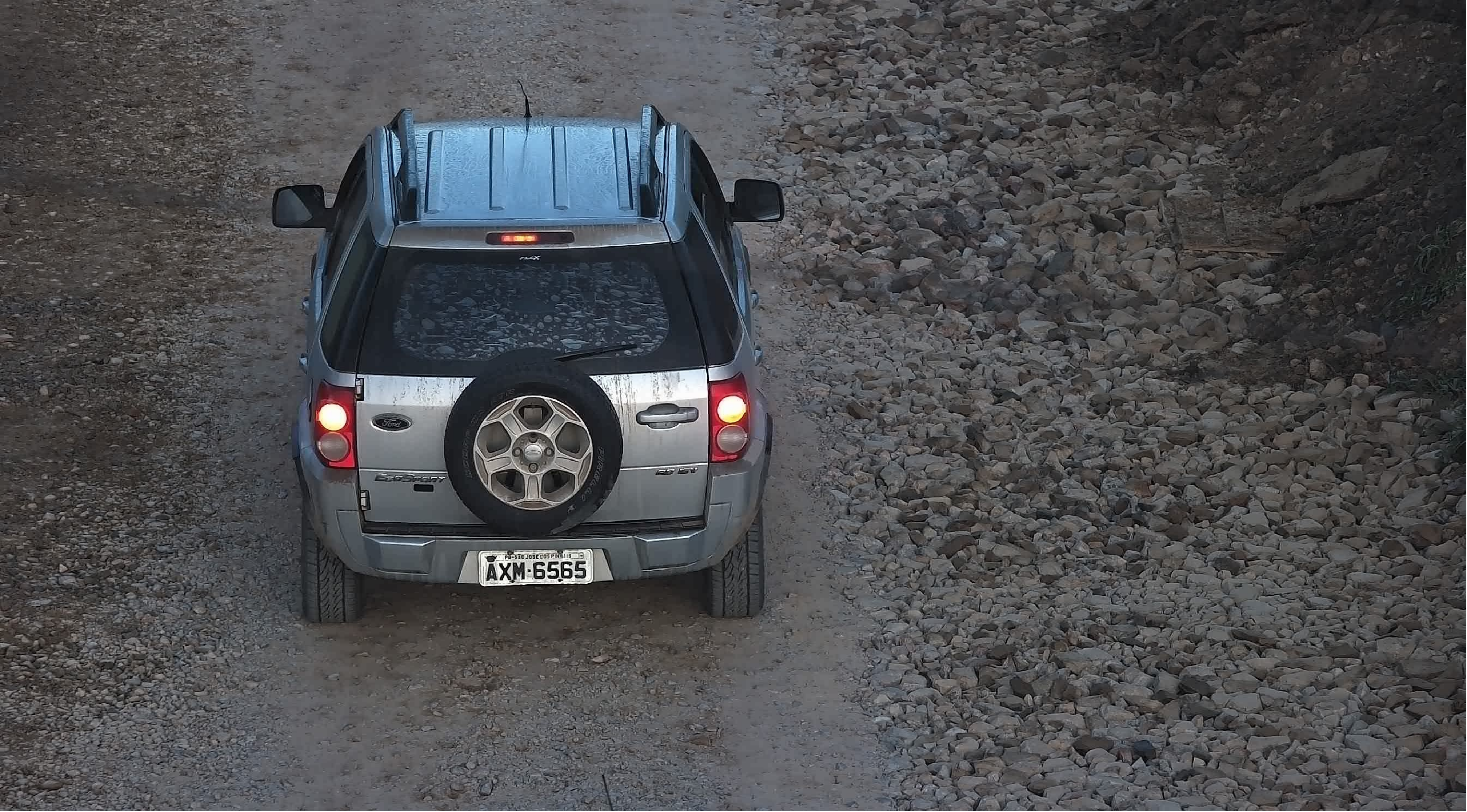}
\includegraphics[height=0.1175\textwidth]{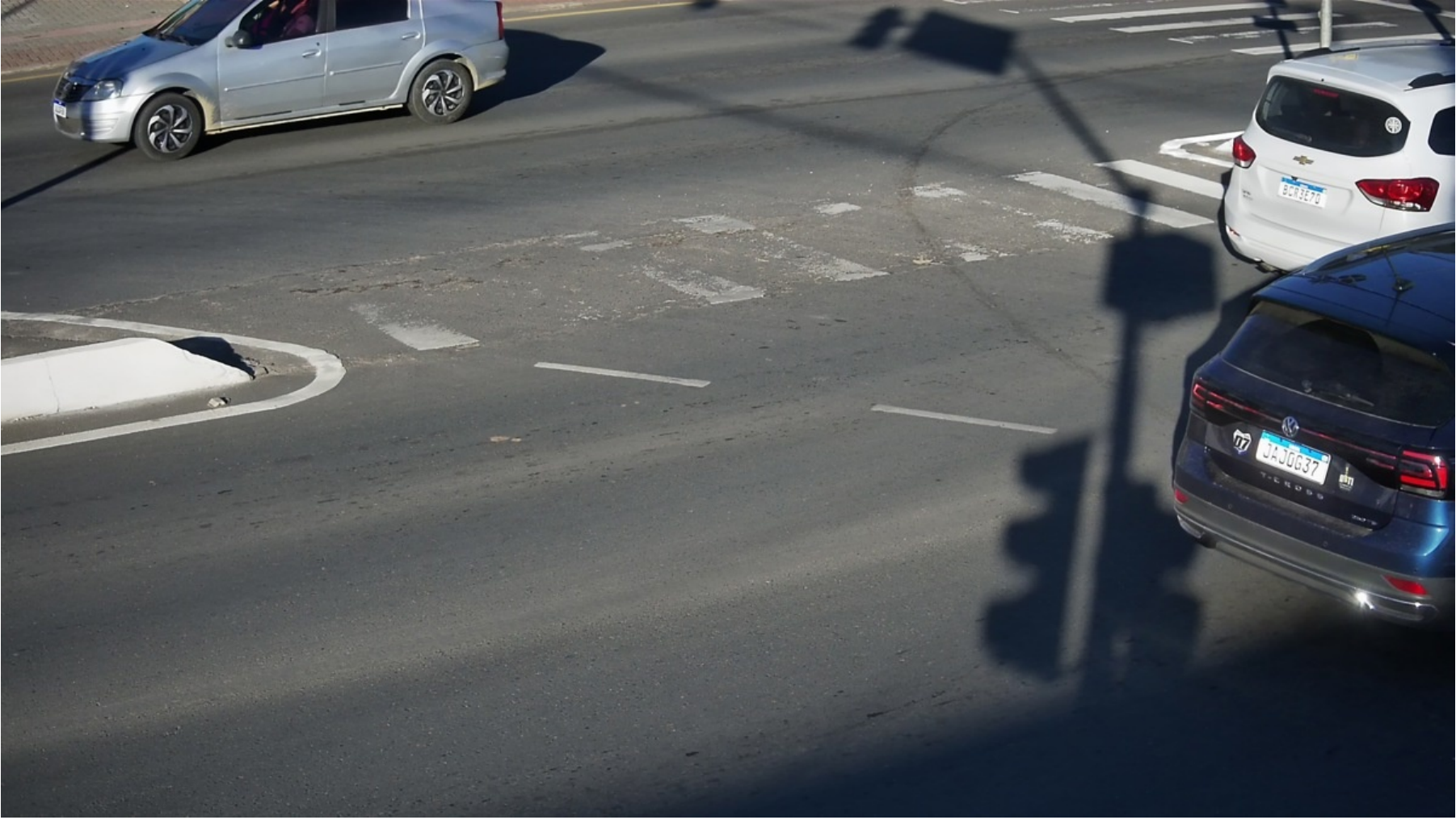}
\includegraphics[height=0.1175\textwidth]{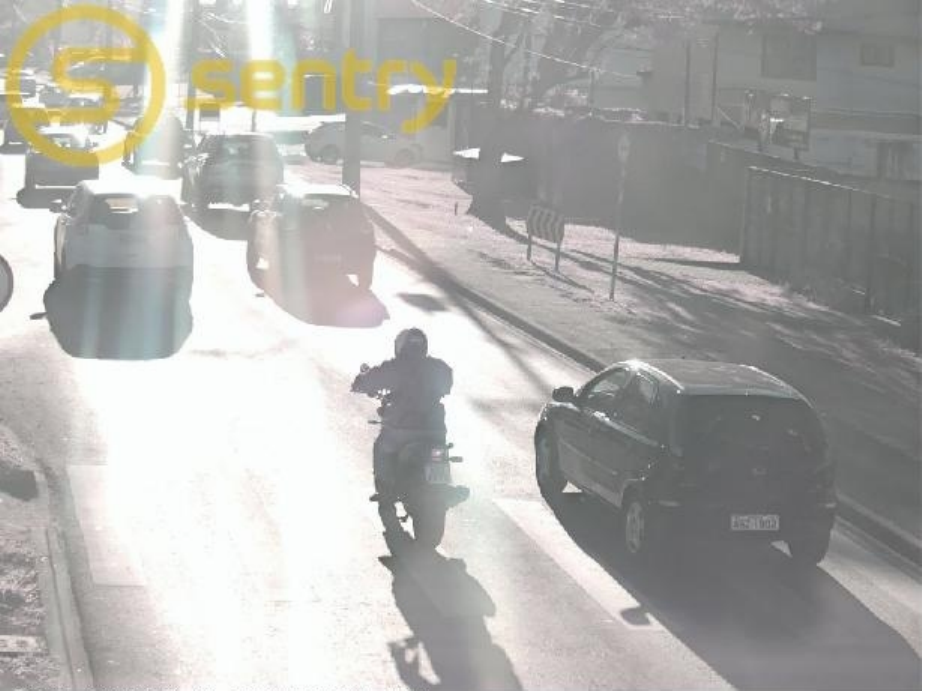}
}
    
\vspace{-2mm}

\caption{Representative samples from the proposed \dataset dataset.
}
\label{fig:lplc_samples}
\vspace{-0.75mm}
\end{figure*}

\section{The proposed Dataset}
\label{sec:dataset}

The proposed dataset\footnote{\hspace{0.2mm}\dataset is available at \urlDataset}, \dataset, is an expanded and enriched version of \previous~\cite{wojcik2025lplc}.
The original benchmark was designed primarily for legibility classification, improving \gls{alpr} pipelines by enabling different processing strategies according to \gls*{lp} legibility.
In particular, distinguishing between legible and illegible \glspl*{lp} allows unsuitable detections to be filtered out prior to \gls*{ocr}. Building upon this foundation, \dataset substantially broadens the scope of the benchmark by incorporating additional annotations relevant to practical scenarios involving legibility assessment, while also significantly increasing the dataset size.

\dataset features new images collected from the same real-world domain as \previous.
These were captured over three different days by traffic radar systems deployed across the Brazilian state of \StateName and were pre-processed prior to release to remove embedded metadata and anonymize capture locations.
Each image contains at least one \gls*{lp}, and is annotated with respect to (i)~the time of day (night, morning, afternoon and evening, each a period of 6 hours and starting at midnight), (ii)~the day and (iii)~camera IDs (with some omitted when unique camera identification was not possible), and two capture-condition flags indicating (iv)~rain and (v)~faulty camera operation (some cameras were observed to produce images with a corrupted red channel, resulting in a characteristic magenta color cast across the entire~image).

For each annotated \gls*{lp}, the dataset provides annotations for (i)~the enclosing rectangular bounding box~(\textit{x}, \textit{y}, \textit{w}, \textit{h}), (ii)~the \gls*{lp} text, (iii)~the legibility level, (iv)~vehicle attributes (make, model, type, and color), and (v)~vehicle occlusion~(boolean). These annotations were initially generated automatically using task-specific models, namely a YOLOv11 detector~\cite{ultralytics2025yolov11} for \gls{lp}  localization, a \parseq model~\cite{bautista2022scene} for \gls*{ocr}, and a ResNet-50 model~\cite{he2016deep} for legibility estimation, and were subsequently manually reviewed to ensure consistency and accuracy.
Vehicle attributes were retrieved from the \gls*{senatran} database based on the recognized \gls*{lp} text. Not all \glspl*{lp} include complete annotations: vehicle attributes are omitted when unavailable for a given \gls*{lp} text, while \gls*{lp} characters are excluded when the \gls*{lp} is too degraded to support reliable human validation.
In some cases, however, otherwise illegible \glspl*{lp} could still be validated using the zoom feature of some of the cameras in the dataset.

\cref{fig:lplc_samples} presents representative samples from \dataset, highlighting its diversity and practical relevance.
The bottom-left image illustrates a capture from a faulty camera, a scenario that, to the best of our knowledge, is not represented in existing public datasets.
The top-right image depicts rainy conditions, while the top-left image shows an \gls*{lp} that would otherwise be illegible, but whose characters could be validated due to the camera zoom.
Together, these examples further highlight the variability present in \dataset, including extreme lighting, different camera distances, heterogeneous capture devices, and diverse vehicle types.
As a result, our dataset offers a faithful and challenging representation of real-world \gls{alpr} operating~conditions.

\cref{tab:image_stats} presents the image statistics of \dataset, including all instances inherited from \previous. In total, $26{,}889$ new images containing $28{,}800$ annotated \glspl*{lp} were added to the original $10{,}210$ images and $12{,}687$ \glspl*{lp} from \previous. As previously stated, the dataset spans three distinct capture days. All images from \previous originate from a single day, with new images comprising the two other additional days: $15{,}284$ images from one day and $11{,}605$ from the other.
Images from the \textit{evening} and \textit{night} correspond to night-time instances (6 p.m. to midnight for evening and midnight to 6 a.m. for night), while \textit{morning} and \textit{afternoon} images correspond to daytime (6 a.m. to noon for morning and noon to 6 p.m. for~afternoon).

\begin{table}[!htb]
\renewcommand{\arraystretch}{1.05}
\caption{\dataset Image Statistics.
}
\label{tab:image_stats}

\vspace{-2mm}

\centering
\begin{tabular}{lclc}
\toprule
\multicolumn{2}{c}{Images by Time of Day} & \multicolumn{2}{c}{Images by Attributes} \\
\cmidrule(r){1-2} \cmidrule(l){3-4}
Class & Number & Attribute & Number \\
\midrule %
Morning   & $10{,}998$  & Faulty Camera & $\phantom{0}3{,}690$ \\
Afternoon & $12{,}799$  & Raining       & $\phantom{00,}770$ \\
Evening   & $\phantom{0}9{,}157$  & Has Camera ID   & $29{,}965$ \\
Night     & $\phantom{0}4{,}145$  & $\rightarrow$ Total Images     & $37{,}099$ \\
\bottomrule
\end{tabular}
\end{table}

\cref{tab:plate_stats} reports analogous statistics for the annotated \glspl*{lp} in \dataset.
All annotations were manually reviewed, and \gls*{ocr} outputs were cross-validated using vehicle data.
Partially occluded \glspl*{lp} were removed from the dataset, alongside the corresponding \gls{lp}-level occlusion labels.
Vehicle-level occlusion is present when vehicle attributes cannot be reliably identified from the full image.

Legibility classes follow the same criteria as in \previous. Perfect \glspl*{lp} exhibit clear characters with no visible distortion. Good \glspl*{lp} contain minor, non-disruptive distortions while remaining fully legible. Poor \glspl*{lp} display more severe distortions but still allow unambiguous character identification. Illegible \glspl*{lp} present substantial degradation that prevents confident character validation, except in cases where camera zoom enables reliable interpretation.

\begin{table}[!htb]
    \renewcommand{\arraystretch}{1.05}
    \caption{\dataset \gls{lp} Statistics.}
    \label{tab:plate_stats}

    \vspace{-2mm}
    
    \centering
    \begin{tabular}{lclcc}
        \toprule
        \multicolumn{2}{c}{LPs by Legibility} & \multicolumn{2}{c}{Other Attributes} \\
        \cmidrule(lr){1-2} \cmidrule(lr){3-4}
        Class & Number & Attribute & Number \\
        \midrule
        Perfect    & $18{,}425$ & Vehicle Visible   & $38{,}089$  \\
        Good       & $10{,}180$ & Vehicle Available & $25{,}506$  \\
        Poor       & $\phantom{0}7{,}520$ & \gls*{lp} Text Available & $36{,}414$  \\
        Illegible  & $\phantom{0}5{,}362$ & $\rightarrow$ Total LPs  & $41{,}487$ \\
        \bottomrule

    \end{tabular}
\end{table}

\begin{figure}[!htb]
    \centering
    \includegraphics[width=0.48\textwidth]{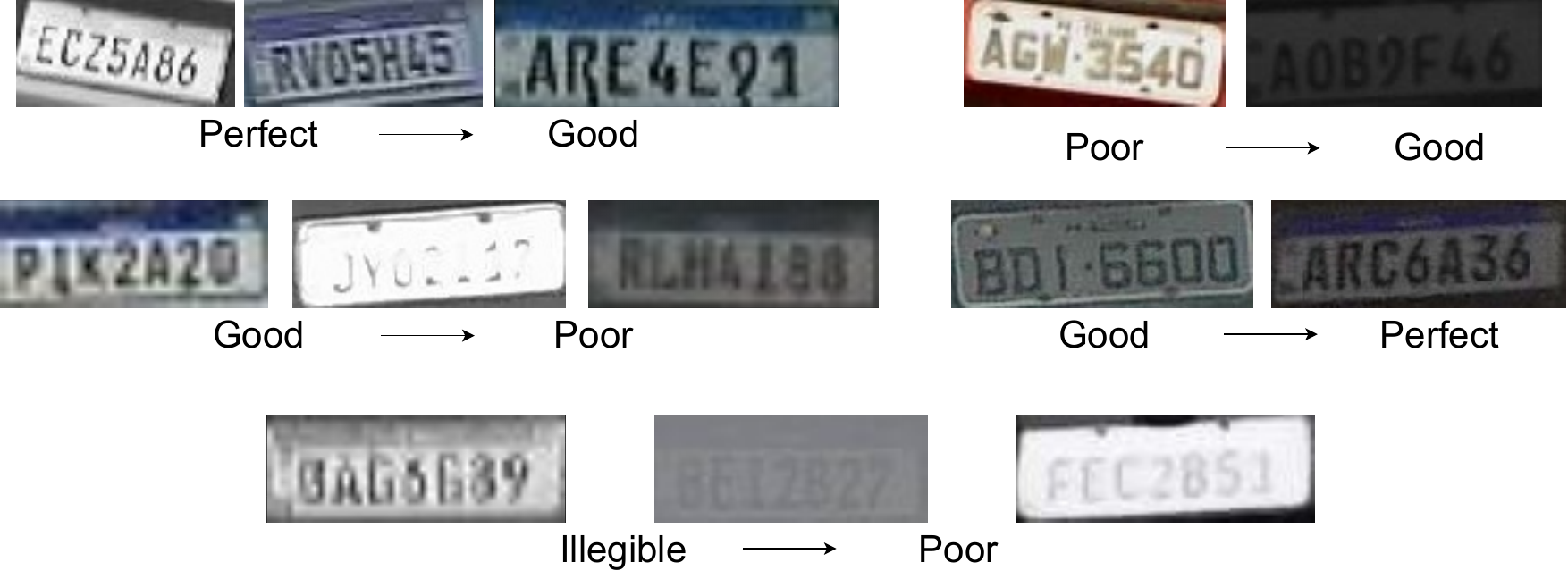}
    \vspace{-2mm}
    \caption{\gls{lp} legibility annotation errors observed in \previous~(left) and their corresponding corrections in \dataset~(right).}
    \label{fig:lp_revision}
\end{figure}

In addition to extending the dataset, we conducted a comprehensive revision of the legibility annotations inherited from \previous.
Several labeling errors, revealed through network mispredictions, were corrected by reassigning affected \glspl*{lp} to adjacent classes. The revision took into account the annotator's assessment together with the predictions from the models used in~\cite{wojcik2025lplc}.
In total, $1{,}012$ \glspl*{lp} were relabeled: $514$ poor instances were reclassified as good, $112$ good instances were reclassified as poor, $275$ perfect instances were reclassified as good, and $111$ good instances were reclassified as perfect.
Furthermore, $787$ \glspl*{lp} originally labeled as illegible were reclassified as poor after successful \gls*{ocr} validation.
Representative examples of these corrections are illustrated in~\cref{fig:lp_revision}.

%% file: 4-setup.tex
\section{Modeling and Experiments}
\label{sec:experiments}

This section presents the experimental setup and modeling decisions adopted to ensure a fair, reproducible evaluation of \gls*{lp} legibility classification across multiple~scenarios.

\subsection{Legibility Classification}

Following~\cite{wojcik2025lplc}, we adopt a $5$-bin, $10$-fold cross-validation protocol with $40$-$20$-$40$ splits for training, validation, and testing. All reported results are the average across the $10$ test folds. To construct the folds, all \glspl*{lp} are uniformly divided into five ordered bins, each containing approximately $20\%$ of the dataset. For each fold, two bins are assigned to training, one to validation, and the remaining two to testing.
The validation bin is circularly rotated across folds~(e.g., the third bin is used for validation in the first fold, the fourth in the second, and so on), producing five distinct splits.
Five additional folds are obtained by swapping the training and testing partitions of each split, resulting in a total of $10$~folds.
All folds are released alongside \dataset to ensure~reproducibility.

When training on \previous~\cite{wojcik2025lplc}, we use the original splits provided with the dataset for a fair comparison with prior results. For \dataset, newly added instances are incorporated exclusively into the training partitions of each fold derived from \previous. As a result, the training set size increases from approximately $5$k to over $33$k \glspl*{lp}.
Naturally, our new experiments utilize the corrected labels for samples from~\previous.

Legibility classification is evaluated under four scenarios. These are classification tasks over cropped \glspl*{lp}.
The first three are reproduced from~\cite{wojcik2025lplc}.
In the ``Baseline'' scenario, the four original legibility classes are used as prediction targets. In the second scenario, denoted ``Legibility Recognition'', the \textit{perfect} and \textit{good} classes are merged into a single label, referred to as \textit{suitable}, and samples labeled as \textit{illegible} are excluded. This label mapping is maintained in the third scenario, ``Full Recognition'', where the \textit{illegible} class is~reintroduced.

The notion of \textit{suitable} is grounded on the reliability of \glspl*{lp} for \gls*{ocr}. Prior results~\cite{wojcik2025lplc} showed that \textit{poor} \glspl*{lp} achieved a whole-plate accuracy of only $72\%$ even with the best-performing model, whereas \textit{good} and \textit{perfect} samples exceeded $90\%$.
Motivated by this performance gap, we introduce a fourth scenario, termed ``Quality Filter'', targeting practical OCR pre-processing.
It is formulated as a binary classification task, where \textit{suitable} comprises the \textit{perfect} and \textit{good} classes, and \textit{unsuitable} comprises the \textit{poor} and \textit{illegible} classes.
\cref{tab:scenarios} summarizes the mappings for each evaluation~scenario.

\begin{table}[!htb]
    \renewcommand{\arraystretch}{1.15}
    \caption{Mapping of original legibility classes to labels used in each evaluation scenario.
    }
    
    \vspace{-2mm}
    \centering
    
    \begin{tabular}{lcccc}
        \toprule
        \textbf{Scenario} &
       
        \multicolumn{4}{c}{\textbf{Classes}} \\

        \midrule
        Baseline &
        Perfect &
        Good &
        Poor &
        Illegible \\

        Legibility Recognition &
        \multicolumn{2}{c}{Suitable} &
        Poor &
        -- \\

        Full Recognition &
        \multicolumn{2}{c}{Suitable} &
        Poor &
        Illegible \\

        Quality Filter &
        \multicolumn{2}{c}{Suitable} &
        \multicolumn{2}{c}{Unsuitable} \\
        \bottomrule
    \end{tabular}
    \label{tab:scenarios}
\end{table}

In addition, we also introduce two new $10$-fold cross-validation splits constructed exclusively from \glspl*{lp} extracted from images with an associated camera identifier. These comprise $33{,}977$ \glspl*{lp} and define the \emph{intra-camera} and \emph{cross-camera} evaluation protocols. The goal is to investigate potential biases in legibility classification arising from testing on data captured by the same devices observed during~training.

The inclusion of camera identifiers for samples originating from \previous reveals that the previously adopted $10$-fold splits exhibited camera-level overlap between training and test sets.
This form of contamination can lead to overly optimistic performance estimates~\cite{laroca2022first}.
We therefore release these camera-aware splits as a new component of the benchmark.

To construct the camera-based splits, the cameras are sorted in descending order according to the number of available images. Two assignment strategies are then applied in parallel. In the first strategy, images from each camera are uniformly distributed across one set of five bins.
In the second strategy, all images from a given camera are assigned to the bin that currently contains the fewest samples within a separate set of five bins.
Both strategies aim to balance the total number of images per bin. Each set of bins is subsequently used to generate a distinct $10$-fold split, corresponding to the intra-camera and cross-camera protocols, respectively.

\subsection{Model and Parameters}

In this work, we use the pre-trained ResNet-50 model~\cite{he2016deep} available in the PyTorch \texttt{torchvision} library.
Although this model was not the top-performing architecture in~\cite{wojcik2025lplc}, its performance improves substantially after the proposed training refinements and the inclusion of additional training~data.

To better handle hard examples, we adopt an \gls*{ema} loss~\cite{morales2024ema}, implemented as a dynamically weighted cross-entropy loss for multi-class classification. The class weights are computed at each mini-batch based on the relative number of misclassified samples per class. The class weight tensor for batch $b$ is defined in \cref{eq:ema_weights}, where $mispred$ corresponds to the error count in the smoothed confusion matrix, as defined in \cref{eq:smooth} and \cref{eq:mispred}. This calculation ensures a higher weight for classes with higher error counts, and the loss is stabilized by using the smoothed confusion matrix for weight calculation. This modeling is suitable for LPLCv2, which features class imbalance and critical differences between the least represented classes.

\begin{equation}
\label{eq:ema_weights}
W(b) = min(\frac{max(mispred(b))}{mispred(b)+\epsilon}, M) \\  
\end{equation}

\begin{equation}
\label{eq:smooth}
smooth(b) = \alpha*confusion(b) + (1-\alpha)*confusion(b) \\  
\end{equation}

\begin{equation}
\label{eq:mispred}
mispred(b) = sum(smooth(b)) - diag(smooth(b))
\end{equation}

Our implementation uses \(\alpha=0.8\), \(\epsilon=1e-6\) and \(M=1.2\).
Optimization is performed using Adam~\cite{kingma2015adam} with an initial learning rate of $1e{-5}$.
A linear decay factor of $0.75$ is applied with a patience of $5$ epochs, down to a minimum learning rate of $5e{-6}$.
Models are trained independently for each fold using early stopping with a patience of $20$ epochs based on validation accuracy, for a maximum of $1{,}000$ epochs. The batch size is set to $64$, and all layers of the network are fine-tuned rather than freezing the backbone.
All experiments are conducted on an NVIDIA RTX 6000~GPU with 48 GB of~RAM.

%% file: 5-results.tex
\section{Results}
\label{sec:results}

\cref{tab:scen_base} reports the results for the \textit{Baseline} scenario.
All results are reported as F1-scores computed on the test sets and averaged across the $10$ folds.
The \emph{Overall} column corresponds to the micro-averaged F1-score across all classes. The first row reproduces the results reported in~\cite{wojcik2025lplc} (\previous). 
Subsequent rows show the cumulative impact of the improvements introduced in this work, namely the correction of legibility labels, the addition of newly annotated images to the training sets, and the usage of the \gls{ema} loss.
The test sets are unchanged, and the model is trained using the same hyperparameters as in the original benchmark unless stated otherwise.

\begin{table}[!htb]
    \centering
    \renewcommand{\arraystretch}{1.05}
    \caption{Baseline results using ResNet-50, reported as F1-score~(\%).
    }
    
    \vspace{-2mm}

    \resizebox{0.99\linewidth}{!}{

    \begin{tabular}{lccccc}
        \toprule
        \multirow{2}[2]{*}{Partition} & \multicolumn{4}{c}{Class} & \multirow{2}[2]{*}{Overall} \\
        \cmidrule(lr){2-5}
        & Perfect & Good & Poor & Illegible & \\
        \midrule
        \previous     & $84.5$\% & $68.0$\% & $56.7$\% & $73.0$\% & $74.5$\% \\
        + Revised  Labels & $92.1$\% & $83.1$\% & $83.8$\% & $88.2$\% & $84.4$\% \\
        + More Images     & $92.8$\% & $86.0$\% & $86.0$\% & $92.9$\% & $88.7$\% \\
        + \gls{ema} & $92.7$\% & $86.2$\% & $86.3$\% & $93.0$\% & $89.8$\% \\ %
        \bottomrule
    \end{tabular}

    }
    \label{tab:scen_base}
\end{table}

The results show that the label correction substantially improves dataset consistency and, consequently, model performance, with the overall F1-score increasing from $74.51$\% to $84.44$\%. Expanding the training data with new samples leads to further improvement, raising performance to $88.67$\%, corresponding to a reduction of approximately $27$\% in the remaining classification errors.
The introduction of the \gls{ema} loss produces an additional gain in overall performance.
In particular, it the average F1-score for the three minority classes~(good, poor and illegible) are slightly improved, with a marginal reduction for the majority class~(perfect).
Upon aggregation, the trade-off results in a higher overall~F1-score.

\begin{figure}[!htb]
    \centering

    \hspace{-1.5mm}\includegraphics[width=0.6\linewidth]{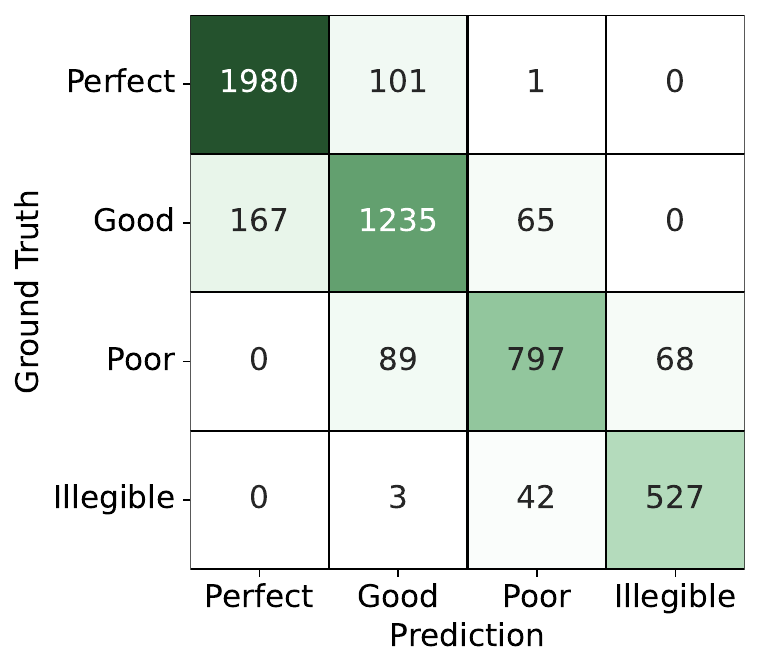}

    \vspace{-3.75mm}
    
    \caption{Confusion matrix from one of the EMA experiment folds.
    }
    \label{fig:cm}
\end{figure}

For the \gls{ema} loss experiment, \cref{fig:cm} presents the confusion matrix for one representative (median F1-score) fold. Most errors occur between adjacent legibility classes, indicating that the model rarely produces extreme misclassifications. For example, \textit{good} \glspl*{lp} are primarily confused with \textit{perfect} or \textit{poor} samples, and analogous patterns are observed for other classes.
This behavior suggests that the learned decision boundaries largely respect the ordinal nature of the legibility~labels.

\begin{figure}[!htb]
    \centering

    \hspace{-1.5mm}\includegraphics[width=0.95\linewidth]{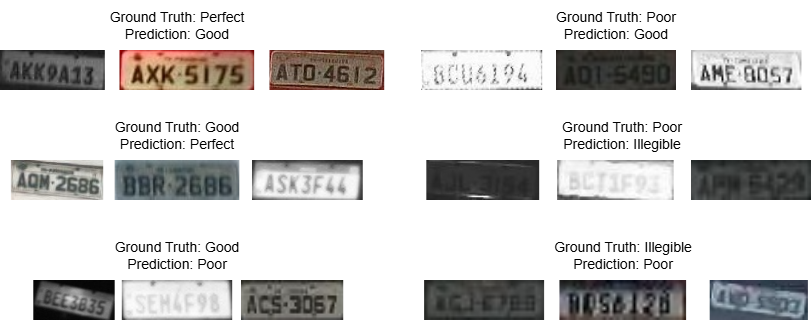}

    \vspace{-2mm}
    
    \caption{Examples of misclassified \glspl*{lp} illustrating borderline legibility cases.
    }
    \label{fig:lp_errors}
\end{figure}

\cref{fig:lp_errors} provides qualitative insight into these misclassifications. Many errors arise from visually ambiguous cases in which the appearance of an \gls{lp} lies close to the boundary between classes.
Some \textit{illegible} \glspl*{lp} retain a small number of recognizable but severely degraded characters, while some \textit{poor} \glspl*{lp} exhibit strong blur patterns similar to those observed in fully illegible instances.
Likewise, certain \textit{good} and \textit{perfect} \glspl*{lp} share comparable noise characteristics, reducing the visual contrast between these categories. These examples highlight that the remaining errors are largely driven by intrinsic ambiguity in the data rather than systematic model~failures.

\begin{figure}[!htb]
    \centering
    \captionsetup[subfigure]{captionskip=1.75pt,font=footnotesize}

    \subfloat[][\textit{Illegible} \glspl*{lp} classified as \textit{good}.]{%
            \includegraphics[height=7ex]{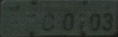}
            \includegraphics[height=7ex]{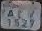}
            \includegraphics[height=7ex]{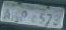}
    }

    \vspace{2mm}

    \subfloat[][\textit{Perfect} \gls*{lp} classified as \textit{poor}.]{%
        \parbox{0.85\linewidth}{%
            \centering
            \includegraphics[height=7ex]{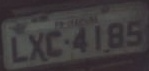}
        }%
    }

    \caption{\glspl*{lp} misclassified into non-adjacent legibility classes.}
    \label{fig:lp_other_errors}
\end{figure}

We further analyze the edge cases in \cref{fig:lp_other_errors}, which correspond to misclassifications where an \gls*{lp} is assigned to a non-adjacent legibility class.
Three \textit{illegible} \glspl*{lp} are incorrectly classified as \textit{good}.
As the current model operates on the entire \gls*{lp} patch rather than on individual characters, we conjecture that background patterns in these samples resemble those of certain \textit{good} instances observed during training.
Conversely, the single \textit{perfect} \gls*{lp} misclassified as \textit{poor} appears to be affected by low-contrast imaging conditions, which are also characteristic of some \textit{poor} samples. Unlike genuinely \textit{poor} \glspl*{lp}, however, the characters in this instance do not exhibit the obtrusive degradation associated with that~class.

We report the results for the remaining scenarios in \cref{tab:other_scens} (see \cref{tab:scenarios} for class mapping).
Performance is measured using the mean micro-averaged F1-score on the test sets, corresponding to the \textit{Overall} metric reported in \cref{tab:scen_base}.
For the \textit{Quality Filter} scenario, not present in LPLCv1~\cite{wojcik2025lplc}, we reproduce the same setup using the original labels and parameters for a fair comparison with the proposed~contributions.

\begin{table}[!htb]
    \centering
    \renewcommand{\arraystretch}{1.05}
    \caption{ResNet-50 performance (F1-score) in the remaining scenarios: Legibility Recognition (Suitable and Poor); Full Recognition (Suitable, Poor and Illegible); and Quality Filter (Suitable and~Unsuitable).} %
    
    \vspace{-2mm}

    \begin{tabular}{lccc}
        \toprule
        \multirow{2}[2]{*}{Partition} & Legibility & Full &  Quality  \\
        & Recognition & Recognition & Filter\\
        \midrule 
        \previous     & $92.6$\% & $87.2$\% & $93.3$\% \\
        + Revised  Labels & $95.3$\% & $93.7$\% & $96.4$\% \\
        + More Images     & $95.9$\% & $95.2$\% & $96.9$\% \\
        + \gls{ema} & $96.5$\% & $94.8$\% & $96.7$\% \\
        \bottomrule
    \end{tabular}
    \label{tab:other_scens}
\end{table}

The results for the corrected labels and expanded dataset follow the same trend observed in the \textit{Baseline} scenario, yielding consistent improvements in the test performance. In contrast, the effect of the \gls{ema} loss is less stable in scenarios with fewer target classes. While performance improves in the \textit{Legibility Recognition} scenario, it decreases slightly in the other two scenarios. Most importantly, in the \textit{Quality Filter} scenario, the variation is minimal and not statistically~significant.

Finally, we perform the camera contamination experiments using the proposed novel protocol.
In \cref{tab:cam_ablation_base} we present the results of the intra-camera and cross-camera experiments conducted under the \textit{Baseline} scenario.
Both experiments use the same hyperparameters (with \gls{ema}), differing only in the adopted $10$-fold splits.
Unlike the previous experiments, where newly annotated samples are added exclusively to the training sets, the camera-based splits are generated over the entire revised benchmark. Consequently, the training partitions are smaller, while the validation and test sets are larger. Even under these more restrictive conditions, the observed performance degradation is limited, indicating strong generalization across camera devices and reinforcing the robustness of the proposed benchmark.

\begin{table}[!htb]
    \centering
    \renewcommand{\arraystretch}{1.05}
    \caption{Baseline scenario results under intra-camera and cross-camera protocols~(F1-score).}
    
    \vspace{-2mm}
    
    \begin{tabular}{lccccc}
        \toprule
        \multirow{2}[2]{*}{Partition} & \multicolumn{4}{c}{Class} & \multirow{2}[2]{*}{Overall} \\
        \cmidrule(lr){2-5}
        & Perfect & Good & Poor & Illegible & \\
        \midrule 
        Intra-Camera    & $93.4$\% & $82.6$\% & $84.8$\% & $91.6$\% & $89.2$\% \\
        Cross-Camera    & $93.5$\% & $81.8$\% & $85.7$\% & $91.1$\% & $88.2$\% \\
        \bottomrule
    \end{tabular}
    \label{tab:cam_ablation_base}
\end{table}

As shown by the results, separating images captured by the same camera across different dataset partitions leads to a measurable but moderate decrease in performance. This indicates that the proposed legibility classification approach is largely robust to the camera variability present in the revised benchmark. The same trend is observed across the remaining scenarios, as reported in \cref{tab:cam_ablation_others}.

\begin{table}[!htb]
    \centering
    \renewcommand{\arraystretch}{1.05}
    \caption{ResNet-50 performance across the other evaluation scenarios under intra-camera and cross-camera splits~(F1-score).}
    
    \vspace{-2mm}
    
    \begin{tabular}{lccc}
        \toprule
        \multirow{2}[2]{*}{Partition} & Legibility & Full &  Quality  \\
        & Recognition & Recognition & Filter\\
        \midrule 
        Intra-Camera    & $96.2$\% & $94.5$\% & $96.6$\% \\
        Cross-Camera    & $95.6$\% & $93.8$\% & $96.2$\% \\
        \bottomrule
    \end{tabular}
    \label{tab:cam_ablation_others}
\end{table}

%% file: 6-conclusions.tex
\section{Conclusions}
\label{sec:conclusion}

This paper advances the study of \gls*{lp} legibility classification by revising a previously proposed benchmark, introducing a substantially expanded dataset, and improving recognition performance through a refined training strategy. We show that correcting annotation errors in the original benchmark improves dataset consistency and leads to significant performance gains. Additional improvements are achieved by expanding the training data and adopting a tailored loss function designed to better handle difficult and ambiguous samples.

Beyond performance gains, we enhance the dataset in both diversity and descriptive power. The proposed dataset incorporates new capture conditions, including rainy environments and imagery from faulty cameras.
To the best of our knowledge, the latter represents the first public contribution of its kind within the broader \gls{alpr} literature.
We also introduce richer annotations, such as camera identifiers for most images and vehicle-related metadata for most \glspl*{lp}.
These additions enable a wider range of experimental protocols and facilitate novel studies to be~conducted.

A viable direction for future work is to integrate legibility classification into a complete \gls{alpr} pipeline by leveraging the dataset’s detailed annotations.
This will likely require new evaluation protocols for fair model comparison and may also enable studies on vehicle identification, such as \gls*{lp} super-resolution~\cite{nascimento2025toward} and fine-grained vehicle classification~\cite{lima2024toward}.
Another direction for further research regards the cross-model generalization of the proposed contributions, as well as cross-dataset evaluation for \glspl*{lp} of different layouts.

%% file: 0-acknowledgments.tex
\iffinal

\section*{Acknowledgments}

This study was supported in part by the \textit{Coordenação de Aperfeiçoamento de Pessoal de Nível Superior~(CAPES), Brasil}, under the \textit{Programa de Excelência Acadêmica~(PROEX)}~--~Finance Code $001$; in part by the \textit{Conselho Nacional de Desenvolvimento Científico e Tecnológico~(CNPq)}~(\#~$315409$/$2023$-$1$); and in part by the \textit{Fundação Araucária}~(\#~$078$/$2026$).
\else
\fi